\documentclass[twoside,11pt]{article}
\usepackage{mhStyle2}

\usepackage{times}
\usepackage{epsfig}
\usepackage{graphicx}
\usepackage{amsmath}
\usepackage{amssymb,url,xspace,caption,array,subcaption}
\usepackage{booktabs,balance}
\usepackage[ruled,vlined]{algorithm2e}
\usepackage[usenames,dvipsnames]{xcolor}

\usepackage[pagebackref=false,breaklinks=true,letterpaper=true,colorlinks,bookmarks=false]{hyperref}

\graphicspath{{./}{./figures/}}
\DeclareGraphicsExtensions{.pdf,.jpeg,.png,.jpg}

\makeatletter
\DeclareRobustCommand\onedot{\futurelet\@let@token\@onedot}
\def\@onedot{\ifx\@let@token.\else.\null\fi\xspace}

\def\eg{\emph{e.g}\onedot} 
\def\ie{\emph{i.e}\onedot}

\def\etal{\emph{et al}\onedot}
\makeatother

\def\Vec#1{{\boldsymbol{#1}}}
\def\Mat#1{{\boldsymbol{#1}}}

\def\GRASS#1#2{\mathcal{G}({#1},{#2})}

\def\SPD#1{\mathcal{S}_{++}^{#1}}

\def\cite{\Citep}

\jmlrheading{}{2015}{Harandi \etal}
\ShortHeadings{When VLAD met Hilbert}{Harandi \etal}

\begin{document}

\title{When VLAD met Hilbert}

\author{\name Mehrtash Harandi \email mehrtash.harandi@nicta.com.au \\
       \addr NICTA and Australian National University\\
       Canberra, Australia
       \AND
       \name Mathieu Salzmann \email mathieu.salzmann@nicta.com.au \\
   	   \addr NICTA and Australian National University\\
       Canberra, Australia
       \AND
       \name Fatih Porikli \email fatih.porikli@anu.edu.au \\
   	   \addr NICTA and Australian National University\\
       Canberra, Australia       }

\maketitle

\begin{abstract}

Vectors of Locally Aggregated Descriptors (VLAD) have emerged as powerful image/video representations that compete with or even outperform state-of-the-art approaches on many challenging visual recognition tasks.
In this paper, we address two fundamental limitations of VLAD:  its requirement for the local descriptors to have vector form and its restriction to linear classifiers due to its high-dimensionality.
To this end, we introduce a kernelized version of VLAD. This not only lets us inherently exploit more sophisticated classification schemes, but also enables us to efficiently aggregate non-vector descriptors (\eg, tensors) in the VLAD framework.
Furthermore, we propose three approximate formulations that allow us to accelerate the coding process while still benefiting from the properties of kernel VLAD.
Our experiments demonstrate the effectiveness of our approach at handling manifold-valued data, such as covariance descriptors, on several classification tasks.  Our results also evidence the benefits of our nonlinear VLAD descriptors against the linear ones in Euclidean space using several standard benchmark datasets.

\end{abstract}

\section{Introduction}
\label{sec:intro}

This paper introduces several nonlinear formulations of \emph{Vectors of Locally Aggregated Descriptors} (VLAD) that generalize their use to manifold-valued local descriptors, such as symmetric positive definite (SPD) matrices, and allows them to inherently exploit more sophisticated classification algorithms.
Modern visual recognition techniques typically represent images by aggregating local descriptors, which, compared to image intensities, provide robustness to varying imaging conditions. 
From a historical point of view, this trend was gained momentum by the \emph{Bag-of-Words} (BoW) model~\cite{Sivic_ICCV_2005,Grauman:ICCV:2005,Lazebnik_CVPR_2006}, which had a significant impact on recognition performance. Since then, the notable recent developments include dictionary-based solutions~\cite{Winn_ICCV_2005,Yang_CVPR_2009}, Fisher Vectors (FV)~\cite{Perronnin_CVPR_2007,Perronnin_ECCV_2010}, VLAD~\cite{Jegou_CVPR_2010,Arandjelovic_CVPR_2013} and Convolutional Neural Networks (CNN)~\cite{Krizhevsky_NIPS_2012}.

Among the aforementioned techniques, VLAD stands out for the following reasons:
\begin{itemize}
\vspace{-2mm}
\item VLAD is computed via primitive operations. This makes VLAD extremely attractive when computational complexity is a concern.
\vspace{-2mm}
\item In contrast to CNNs, training a VLAD encoder is straightforward and not contingent on having a large training set.
\vspace{-2mm}
\item VLAD can be considered as a special case of FVs and hence inherits several properties of FVs. The most eminent one is its theoretical connection to the Fisher kernel~\cite{Jaakkola_NIPS_1999}. 
\vspace{-2mm}
\item From an empirical point of view, VLAD has been shown to either deliver state-of-the-art accuracy, or compete with the state-of-the-art methods. For instance, for scene classification on the MIT Indoor dataset, multi-scale VLAD, with only 4096 features, comfortably outperforms the mixture of FV and bag-of-parts, which relies on 221550 features~\cite{Gong_ECCV_2014}.
\end{itemize}

Despite its unique features, VLAD comes with its own limitations. 
In particular, VLAD is designed to work with local descriptors in the form of vectors. Yet, several recent studies in computer vision suggest that structural data (\eg, SPD matrices, graphs, orthogonal matrices) have the potential to provide more robust descriptors. Furthermore, since VLAD typically yields a high-dimensional image representation, it is mostly restricted to employing linear classifiers. Nonetheless, the effectiveness of kernel-based methods has been proven many a time in visual 
recognition~\cite{Gehler_CVPR_2009,Bo_NIPS_2010,Perronnin_CVPR_2010,Vedaldi_TPAMI_2012}.

In this paper, we present kernel based formulations of VLAD to address the  aforementioned shortcomings. In particular, we first introduce a kernelized version of VLAD that relies on mapping of each local descriptor to a Reproducing Kernel Hilbert Space (RKHS). Since several valid kernel functions have recently been defined for non-vector data~\cite{Sadeep:CVPR:2013,Harandi_ECCV_2014}, such a RKHS mapping can be applied to descriptors on different manifold topologies including SPD matrices and linear subspaces (Grassmannian). Having a RKHS mapping, we can aggregate VLAD over various geometries, thus, ultimately generalize the use of VLAD to local descriptors defined in non-vector spaces. Furthermore, the inherent nonlinearity of mappings to RKHS allows us to exploit more advanced classifiers with kernel VLAD.

In the spirit of computational efficiency, we also design three novel nonlinear approximations of our kernel VLAD; a Nystr\"{o}m method that obtains an explicit mapping to the Hilbert space, a local subspace-based representation of the data in Hilbert space, and a Fourier approximations based on the Bochner theorem. These approximations enjoy the similar properties of kernel VLAD, yet have the additional benefit of providing us with faster coding schemes. 
{Interestingly, all our algorithms mostly preserve the simplicity of VLAD, in the sense that the extra computations 
merely consist of kernel evaluations potentially followed by projections (\ie, matrix multiplications).}

Table~\ref{tab:kvlad_sneak_peek} provides a summary of the proposed methods and their attributes.
{
Since each algorithm possesses unique features, it is not truly possible to pick one of the proposed methods as the ultimate winner.
However, our experiments suggest that sVLAD is a good compromise
between speed and accuracy, and would thus be our recommendation.
}


\begin{table}[t]
	\small
	\centering
	\caption
    {\small Proposed methods and their properties. \textbf{Kernel} denotes the type of kernel function the algorithm can accept. For example, the \emph{fVLAD} algorithm can only work with certain type of kernel functions while the \emph{kVLAD} method can accept all type of kernel functions. \textbf{Coding} reflects the form of the output of the algorithm. For example, in the case of \emph{kVLAD}, the output codes are only known implicitly. \textbf{Complexity} is the computational load. 
    }
   \label{tab:kvlad_sneak_peek}
  \begin{tabular}{l c c c}
    \toprule
    {\bf Method}   			&{\bf Kernel} 	&{\bf Coding}					
    &{\bf  Complexity}\\
    \toprule
    \textbf{kVLAD }			& general 		& Implicit	& High	\\
    \textbf{nVLAD }			& general 		& Explicit	& Low	\\
    \textbf{fVLAD }			& specific 		& Explicit	& Low	\\   
    \textbf{sVLAD }			& general 		& Explicit	& Low	\\
	\bottomrule
  \end{tabular}
\end{table}

Our experimental evaluation demonstrate the effectiveness of our approach at handling manifold-valued data in a VLAD framework. Furthermore, we evidence the benefit of exploiting nonlinear classifiers for visual recognition by comparing the performance of our nonlinear VLAD with the standard one on several benchmark datasets, where the local descriptors have a vector form.

\subsection{Related Work}

Most of the popular image classification methods extract local descriptors (\ie, at patch level), 
which are then aggregated into a global image representation~\cite{Lazebnik_CVPR_2006,Perronnin_CVPR_2007,Jegou_CVPR_2010,Perronnin_ECCV_2010,Van_PAMI_2010,Krizhevsky_NIPS_2012,Arandjelovic_CVPR_2013}.

When large amounts of training data are available, CNNs have now emerged as the method of choice to learn local descriptors. With limited number of training samples, existing methods typically opt for handcrafted features, such as SIFT.

To aggregate local features, in addition to operations such as average-pooling and max-pooling,  
histogram-based solutions (\eg, BoW) have proven successful. Going beyond simple histograms has been an active topic of research for the past decade. For instance,~\cite{Lazebnik_CVPR_2006} aggregates histograms computed over different spatial regions. More recent developments, such as FVs~\cite{Perronnin_CVPR_2007} and VLAD~\cite{Jegou_CVPR_2010}, suggest that high-order statistics should be encoded in the aggregation process.

In a separate line of research, structured descriptors (\eg, covariance descriptors or linear subspaces)
have been shown to provide robust visual models~\cite{Tuzel_PAMI_2008,Sadeep:CVPR:2013,Harandi_ICCV_2013}. Being of a non-vectorial form, aggregating such descriptors is hard to achieve beyond simple histograms. Nonetheless, one would like to benefit from the best of both worlds, that is, using robust non-vectorial descriptors in conjunction with state-of-the-art aggregation techniques, such as VLAD. This, in essence, is what we propose to achieve in this paper via kernelization. Furthermore, our approach has the additional advantage of allowing us to inherently exploit nonlinear classifiers that have proven powerful in visual recognition.

While a full review of kernel-based methods in computer vision is beyond the scope of this paper, the recent work of~\cite{Mairal_NIPS_2014} is of particular relevance here.~\cite{Mairal_NIPS_2014} introduces an approach to employing kernels within a CNN framework. Here, we perform a similar analysis within the VLAD framework, with the additional benefit of obtaining a representation that lets us work with manifold-valued data.

\section{Nonlinear VLAD}
\label{sec:kvlad}

In this section, we derive several nonlinear formulations of VLAD. To this end, we first start by reviewing the conventional VLAD and then discuss our approach to kernelizing it, followed by three approximations of the resulting kernel VLAD.

\subsection{Conventional VLAD}
\label{subsec:euclidean_vlad}

Let $\mathcal{X} = \{ \Vec{x}_i \}_{i=1}^N, \Vec{x}_i  \in \mathbb{R}^d$ be a set of local descriptors extracted from a query image or a video.
In VLAD~\cite{Jegou_CVPR_2010}, 
the input space $\mathbb{R}^d$ is partitioned into $m$ Voronoi cells by means of a codebook $\mathcal{C}$ with centers
$\{\Vec{c}_j\}_{j=1}^m,~\Vec{c}_i \in \mathbb{R}^d $. To obtain the codebook, the k-means algorithm is typically employed. Nevertheless,  the use of supervised algorithms has also recently been advocated to build more discriminative codebooks~\cite{Peng_ECCV_2014}. The VLAD code $\Vec{v} \in \mathbb{R}^{md}$ for the query set $\mathcal{X}$ is obtained by concatenating $m$ Local Difference Vectors (LDV) $\delta_j$
storing, for each center, the sum of the differences between this center and each local descriptor assigned to this center. This can be written as
\begin{equation}
	\Vec{v}(\mathcal{X}) = \Big[ \delta_1^T(\mathcal{X}),\delta_2^T(\mathcal{X}),\cdots,\delta_m^T(\mathcal{X}) \Big]^T\;,
	\label{eqn:euc_vlad_eqn0}
\end{equation}
where

\begin{equation}
	\delta_j(\mathcal{X}) = \sum_{i=1}^N  a_{j}^i\big(\Vec{c}_j - \Vec{x}_i \big)\;,
	\label{eqn:euc_vlad_eqn1}
\end{equation}
with $a_{j}^i$ a binary weight encoding whether the local descriptor $\Vec{x}_i$ belongs to the Voronoi cell with center $\Vec{c}_j$ or not, 
\ie, $a_{j}^i = 1$ if and only if the closest codeword to $\Vec{x}_i$ is $\Vec{c}_j$.

\subsection{Kernel VLAD (kVLAD)}
\label{subsec:kvlad}

As mentioned earlier, the conventional VLAD is designed to work with local descriptors of a vectorial form. As such, it cannot handle structured data representations, such as SPD matrices, or subspaces. While such representations could in principle be vectorized, this would (i) yield impractically high-dimensional VLAD vectors; and (ii) ignore the geometry of these structured representations, which has been demonstrated to result in accuracy losses~\cite{Pennec_IJCV_2006,Tuzel_ECCV_2006,Tuzel_PAMI_2008,Sadeep:CVPR:2013}. Here, we propose to address this problem by kernelizing VLAD.

To this end, let us redefine the query set of local descriptors as $\mathcal{X} = \{ \Vec{x}_i \}_{i=1}^N, \Vec{x}_i  \in \mathbb{X}$, where each descriptor lies in the space $\mathbb{X}$, which, in contrast to VLAD, is not restricted to be $\mathbb{R}^d$. In fact, the only constraint we impose is that $\mathbb{X}$ comes with a valid positive definite \emph{pd} kernel $k: \mathbb{X} \times \mathbb{X} \to \mathbb{R}$. For example, $\mathbb{X}$ could be the space of SPD matrices, with the Gaussian kernel defined in~\cite{Sra_NIPS_2012,Sadeep:CVPR:2013}. 
According to the Moore-Aronszajn Theorem~\cite{Aronszajn_1950}, a \emph{pd} kernel $k(\cdot,\cdot)$ induces a unique Hilbert space on $\mathbb{X}$, denoted hereafter by $\mathcal{H}$, with the property that there exists a mapping $\phi:\mathbb{X} \to \mathcal{H}$, such that $k(\Vec{x},\Vec{y}) = 
\langle \phi(\Vec{x}),\phi(\Vec{y})\rangle_\mathcal{H} = \phi(\Vec{x})^T\phi(\Vec{y})$. Here, we propose to make use of this property to map the local descriptors to $\mathcal{H}$, which is a vector space, and perform a VLAD-like aggregation in Hilbert space. The main difficulty arises from the fact that $\mathcal{H}$ may be infinite-dimensional, and, more importantly, that the mapping $\phi$ corresponding to a given kernel $k$ is typically unknown.

Let us suppose that we are given a codebook $\mathcal{C} = \{ \phi(\Vec{c}_i)\}_{i = 1}^{m}$ in $\mathcal{H}$. For instance, this codebook can be computed using kernel kmeans.
To compute a VLAD code in $\mathcal{H}$, we need to provide solutions for the following operations:
\begin{enumerate}
\item Determine the assignments $\{a_{j}^i\}$ in $\mathcal{H}$.
\item Express the LDVs in $\mathcal{H}$. 
\end{enumerate}

\noindent
To determine the assignments, we note that 
\begin{equation}
\|\phi(\Vec{x}) - \phi(\Vec{y})\|^2 = k(\Vec{x},\Vec{x}) - 2k(\Vec{x},\Vec{y}) + k(\Vec{y},\Vec{y})\;.
\label{eqn:dist_kernel}
\end{equation}
Therefore, for each local descriptor, the nearest codeword can can be determined using kernel values only, \ie, without having to know the mapping $\phi$, which lets us directly define the assignments.

Unfortunately, expressing the LDVs in $\mathcal{H}$ is not this straightforward. Clearly, the form of the LDVs, given by 
\begin{equation*}
\delta_j(\mathcal{X}) = \sum{ a_{j}^i\Big( \phi(\Vec{c}_j) - \phi(\Vec{x}_i) \Big)}\;,
\end{equation*}
with $a_j^i$ obtained using Eq.~\ref{eqn:dist_kernel}, cannot be computed explicitly if the mapping $\phi$ is unknown, which is typically the case for popular kernels, such as RBF kernels.
However, in most practical applications, the VLAD vector is not important by itself; What really matters for visual recognition is a notion of distance between two VLAD vectors. We therefore turn to the problem of computing the distance of two VLAD vectors in Hilbert space.

To this end, let
$\mathcal{X} = \{ \Vec{x}_i \}_{i=1}^{N_\mathcal{X}}, \Vec{x}_i  \in \mathbb{X}$ and 
$\mathcal{Y} = \{ \Vec{y}_i \}_{i=1}^{N_\mathcal{Y}}, \Vec{y}_i  \in \mathbb{X}$ be two sets of local descriptors. The implicit VLAD code of $\mathcal{X}$ in $\mathcal{H}$ can be expressed as
\begin{equation*}
	\Vec{v}_{\mathcal{H}}(\mathcal{X}) = \Big[ \delta_1^T(\mathcal{X}),\delta_2^T(\mathcal{X}),\cdots,\delta_m^T(\mathcal{X}) \Big]^T\;,
\end{equation*}
and similarly for $\Vec{v}_{\mathcal{H}}(\mathcal{Y})$.
Now, we have
\begin{small}
\begin{align}	
	&\Big \langle \Vec{v}_{\mathcal{H}}(\mathcal{X}) , \Vec{v}_{\mathcal{H}}(\mathcal{Y}) \Big \rangle_\mathcal{H} = 
	\sum_{s = 1}^{m} \delta_s^T(\mathcal{X})\delta_s(\mathcal{Y})\notag\\
	&=\sum_{s = 1}^{m}\limits\sum_{i = 1}^{N_\mathcal{X}}\limits\sum_{j = 1}^{N_\mathcal{Y}}\limits
	a_s^ia_s^j  \Big( \phi(\Vec{c}_s) - \phi(\Vec{x}_i) \Big)^T\Big( \phi(\Vec{c}_s) - \phi(\Vec{y}_j) \Big)\notag\\
	&=\sum_{s = 1}^{m}\limits\sum_{i = 1}^{N_\mathcal{X}}\limits\sum_{j = 1}^{N_\mathcal{Y}}\limits
	a_s^ia_s^j
	\Big( k(\Vec{x}_i,\Vec{y}_j) + k(\Vec{c}_s,\Vec{c}_s) 
	- k(\Vec{x}_i,\Vec{c}_s) - k(\Vec{y}_j,\Vec{c}_s) \Big)\;,
	\label{eqn:nl_vlad_sim_XY}
\end{align}
\end{small}
which again only depends on kernel values.

With this inner product, a linear SVM, in its dual form, can directly be used for classification\footnote{Note that this will yield a slightly different optimization problem than the standard kernel-based formulation, since in our case the inner product itself depends on several kernel values.}. In our experiments, we rely on this approach, which we refer to as kernel VLAD or {\bf kVLAD} for short. This inner product, however, also allows us to employ an RBF-based kernel SVM, since
\begin{align*}
&\|\Vec{v}_{\mathcal{H}}(\mathcal{X}) - \Vec{v}_{\mathcal{H}}(\mathcal{Y})\|^2 = 
\langle \Vec{v}_{\mathcal{H}}(\mathcal{X}) , \Vec{v}_{\mathcal{H}}(\mathcal{X}) \rangle_\mathcal{H} 
- 2 \langle \Vec{v}_{\mathcal{H}}(\mathcal{X}) , \Vec{v}_{\mathcal{H}}(\mathcal{Y}) \rangle_\mathcal{H}
+ \langle \Vec{v}_{\mathcal{H}}(\mathcal{Y}) , \Vec{v}_{\mathcal{H}}(\mathcal{Y}) \rangle_\mathcal{H}\;.
\end{align*} 
Note that this essentially yields two layers of kernels, \ie, the RBF kernel of the SVM makes use of the distance, which itself is expressed in terms of kernel values. 


While effective in practice, our kVLAD algorithm, as any kernel method, becomes computationally expensive when dealing with large datasets. In the remainder of this section, we therefore introduce three approximations to kVLAD, that address this limitation while still benefiting from the nice properties of kVLAD.

\subsection{Nystr\"{o}m Approximation (nVLAD)} 
\label{subsec:nystrom_sol}
As a first approximation to kVLAD, we propose to make use of the Nystr\"{o}m method. Following~\cite{Perronnin_CVPR_2010}, this lets us obtain an explicit form for the mapping $\phi$ to the Hilbert space $\mathcal{H}$, and thus allows us to approximate a given kernel.

More specifically, let $\mathcal{T}=\{\Vec{t}_i\}_{i=1}^{M},~\Vec{t}_i \in \mathbb{X}$ be a collection of $M$ training examples, and let $\Mat{K}$ be the corresponding kernel matrix, \ie, $[\Mat{K}]_{i,j} =  k(\Vec{t}_i , \Vec{t}_j)$. We seek to approximate the elements of $\Mat{K}$ as inner products between $r$-dimensional vectors. In other words, we aim to find a matrix $\Mat{Z} \in \mathbb{R}^{r \times M}$, such that $\Mat{K} \simeq \Mat{Z}^{\small T}\Mat{Z}$. The best such approximation in the east-squares sense is given by $\Mat{Z} = \Mat{\Sigma}^{1/2}\Mat{V}$, with $\Mat{\Sigma}$ and $\Mat{V}$ the top $r$ eigenvalues and corresponding eigenvectors of $\Mat{K}$. 
From the Nystr\"{o}m method, for a new sample $\Vec{x} \in \mathbb{X}$, the $r$-dimensional vector representation of the space induced by $k(\Vec{x},\cdot)$ can be written as
\begin{equation}
z_N(\Vec{x}) = \Mat{\Sigma}^{-1/2}\Mat{V}
\Big[k(\Vec{x},\Vec{t}_1),\cdots,k(\Vec{x},\Vec{t}_M)\Big]^T.
\label{eqn:Nystrom_map}
\end{equation}

Given a set of local descriptors $\mathbb{X} = \{\Vec{x}_i\}$, our {\bf nVLAD} algorithm then consists of computing the corresponding $\{z_N(\Vec{x}_i)\}$, and making use of Eq.~\ref{eqn:euc_vlad_eqn0} and Eq.~\eqref{eqn:euc_vlad_eqn1} with this new representation.

\subsection{Local Subspace Approximation (sVLAD)}
\label{subsec:subspace_kVLAD}

Here, we introduce a novel approximation of the Hilbert space $\mathcal{H}$ based on the idea of local subspaces. To this end, we first note that the Nystr\"{o}m approximation yields one single global estimate of $\mathcal{H}$, used across all the codewords and all the descriptors. However, by looking at Eq.~\eqref{eqn:euc_vlad_eqn1}, we can see that the contribution of each codeword in the VLAD vector is independent of the other codewords, particularly since each local descriptor is assigned to a single codeword. Therefore, there is no reason for the approximation of $\mathcal{H}$ to be shared across all the codewords and descriptors. This motivates us to define approximate spaces for each codeword individually. 

To this end, let  $\{\Vec{t}_{s,j}\}_{j=1}^{N_s}$ be the set of training samples that generate the codeword $\Vec{c}_s$. In other words, as in the conventional VLAD where $\Vec{c}_s = \frac{1}{N_s}\sum_j \Vec{t}_{s,j}$, we have $\phi(\Vec{c}_s) = \frac{1}{N_s}\sum_i \phi(\Vec{t}_{s,j})$. While, due to the unknown nature of $\phi$, such a codeword cannot be explicitly computed, we can still evaluate the kernel function at this codeword, since
\begin{align*}
k(\Vec{x},\Vec{c}_s) &= \phi(\Vec{x})^T \phi(\Vec{c}_s) = \frac{1}{N_s}\sum_j \phi(\Vec{x})^T \phi(\Vec{t}_{s,j}) 
= \frac{1}{N_s}\sum_j k(\Vec{x}, \Vec{t}_{s,j})\;.
\end{align*}
Here, we therefore propose to exploit the subspaces spanned by the training samples associated to each individual codeword to obtain an approximate representation of $\mathcal{H}$.

More specifically, let $\mathcal{S}_s = span( \{\phi(\Vec{t}_{s,j})\}_{j=1}^{N_s} )$. We then define
\begin{align}
\overline{\delta}_s(\mathcal{X}) = \sum_{i=1}^N{ a_{s}^i\Big( \pi_s \big( \phi(\Vec{c}_s) \big) - \pi_s \big( \phi(\Vec{x}_i) \big) \Big)}\;,
\label{eqn:ksvlad}
\end{align}
with $\pi_s: \mathcal{H} \to \mathcal{S}_s$ the projection onto $\mathcal{S}_s$. These projections can be obtained following a similar intuition as for nVLAD. More precisely, let $\Mat{K}_s$ be the kernel matrix estimated from the training samples generating $\Vec{c}_s$, \ie, $[\Mat{K}_s]_{i,j} = k(\Vec{t}_{s,i},\Vec{t}_{s,j})$. By eigendecomposition, we can write $\Mat{K}_s = \Mat{U}_s \Lambda_s \Mat{U}_s^T$. Then, $\Phi_s \Mat{U}_s \Lambda_s^{-1/2}$, with $\Phi_s = [\phi(\Vec{t}_{s,1}),\cdots,\phi(\Vec{t}_{s,N_s})]$, forms a basis for $\mathcal{S}_s$. As such, we can write
\begin{equation}
\pi_s(\Vec{x}) = \Lambda_s^{-1/2} \Mat{U}_s \Big[k(\Vec{x},\Vec{t}_{s,1}),\cdots,k(\Vec{x},\Vec{t}_{s,N_s})\Big].
\label{eqn:skVLAD_proj}
\end{equation}
The LDVs $\overline{\delta}_s(\mathcal{X})$ can then be obtained for all codeword $\Vec{c}_s$, and concatenated to for the final {\bf sVLAD} representation.

\begin{remark}
Note that one can also use only the top $r$ eigenvectors of $\Mat{K}_s$ to construct an $r$-dimensional local subspace in $\mathcal{H}$. This would not only yield the same dimensionality for all local subspaces, but could also potentially help discarding the noise associated to the $\{\Vec{t}_{s,i}\}_{i=1}^{N_s}$.
\end{remark}

\subsection{Fourier Approximation (fVLAD)} 
\label{subsec:fourier}

The previous two approximations apply to general kernels and both Euclidean and non-Euclidean data. In the Euclidean case, however, other approximations have been proposed for specific kernels~\cite{Rahimi_NIPS_2007,Vedaldi_PAMI_2012}. Since our experiments on Euclidean data all rely on RBF kernels, here, we discuss an approximation of this type of kernels based on the Bochner Theorem~\cite{Rudin_2011}.

According to the Bochner Theorem~\cite{Rudin_2011}, a shift-invariant kernel\footnote{A kernel function is shift invariant if $k(\Vec{x}_i,\Vec{x}_j) = k(\Vec{x}_i - \Vec{x}_j)$.}, such as Euclidean RBF kernel, can be obtained by the Fourier integral. As shown in~\cite{Rahimi_NIPS_2007}, for real-valued kernels, this can be expressed as
\begin{equation}
k(\Vec{x}_i - \Vec{x}_j) = \int_{\mathbb{R}^d} p(\omega)z_F(\Vec{x}_i)z_F(\Vec{x}_j)d\omega,
\end{equation}
where $z_F(\Vec{x}) = \sqrt{2}\cos(\omega^T\Vec{x}+b)$, with $b$ a random variable drawn from $[0, 2\pi]$. In other words, $k(\Vec{x}_i,\Vec{x}_j) = k(\Vec{x}_i - \Vec{x}_j)$ is the expected value of $z_F(\Vec{x}_i)z_F(\Vec{x}_j)$ under the distribution $p(\omega)$. For the RBF kernel $k(\Vec{x}_i,\Vec{x}_j) = \exp(-\|(\Vec{x}_i - \Vec{x}_j)\|^2/2\sigma^2)$, we have $p(\omega) = \mathcal{N}(0,\sigma^{-2}\mathbf{I}_{d})$. 

Let $\{\omega_i\}_{i = 1}^r$, $\omega_i \in \mathbb{R}^d$, be i.i.d.  samples drawn form the normal distribution $\mathcal{N}(0,\sigma^{-2}\mathbf{I}_{d})$, and $\{b_i\}_{i = 1}^r$ be samples uniformly drawn from $[0,2\pi]$. Then, the $r$ dimensional estimate of $\phi(\Vec{x}) \in \mathcal{H}$ is given by
\begin{equation}
z_F(\Vec{x}) = \sqrt{\frac{2}{r}} \Big[\cos(\omega_1^T \Vec{x}+b_1) , \cdots,\cos(\omega_r^T \Vec{x}+b_r)\Big].
\label{eqn:RFF_map}
\end{equation}

Similarly to nVLAD, we can then compute $z_F(\Vec{x}_i)$ for each local descriptor $\Vec{x}_i$, and use Eq.~\eqref{eqn:euc_vlad_eqn0} and Eq.~\eqref{eqn:euc_vlad_eqn1} to obtain a code. In our experiments, we refer to this approach, which only applies to Euclidean data, as {\bf fVLAD}.

\subsection{Further Considerations}
\label{sec:further_discussions}

\paragraph{Normalization:}\mbox{}\\
Recent developments have suggested that the discriminatory power of VLAD could be boosted by additional post-processing steps, such as $\ell_2$ power normalization and signed square rooting normalization~\cite{Arandjelovic_CVPR_2013,Gong_ECCV_2014}. The $\ell_2$  power normalization, where each block in VLAD is normalized individually, can easily be performed in kVLAD, since
\begin{small}
\begin{align*}
\|\delta_s(\mathbb{X})\|_\mathcal{H}^2 \hspace{-0.5ex} = \hspace{-1.5ex}
\sum_{i,j = 1}^{N_\mathbb{X}}\limits \hspace{-0.75ex} a_s^ia_s^j
	\hspace{-0.25ex} \Big( \hspace{-0.5ex} k(\Vec{x}_i,\Vec{x}_j) \hspace{-0.5ex}  + \hspace{-0.5ex} k(\Vec{c}_s,\Vec{c}_s) 
	\hspace{-0.5ex} - \hspace{-0.5ex} k(\Vec{x}_i,\Vec{c}_s) \hspace{-0.5ex} - \hspace{-0.5ex} k(\Vec{x}_j,\Vec{c}_s)
	\hspace{-0.5ex} \Big) 	
\end{align*}
\end{small}
is only dependent on kernel values. As a result, the inner product of Eq.~\ref{eqn:nl_vlad_sim_XY} after normalizing each VLAD block independently, \ie,
\begin{align*}
\Big \langle \Vec{\bar{v}}_{\mathcal{H}}(\mathbb{X}) , \Vec{\bar{v}}_{\mathcal{H}}(\mathbb{Y}) \Big \rangle_\mathcal{H} =
\sum_{s =1}^{k}\frac{\Big \langle \delta_s(\mathbb{X}),\delta_s(\mathbb{Y})\Big \rangle}
{\|\delta_s(\mathbb{X})\|_\mathcal{H}\|\delta_s(\mathbb{Y})\|_\mathcal{H}}\;,
\end{align*}
will also only depend on kernel values.
By contrast, however, the signed square rooting normalization can only be achieved when explicit forms of the descriptors are available, \ie, in nVLAD, sVLAD and fVLAD.

\paragraph{Kernelizing Fisher Vectors:}\mbox{}\\
Due to the connection between VLAD and FVs, it seems natural to rely on the ideas discussed above to kernerlize FVs.
One difficulty in kernelizing FV, however, arises from the fact that Gaussian distributions, which are required to model the probability distributions in FV, are not well-defined in RKHS. 
More specifically, to fit a Gaussian distribution in a $d$-dimensional space, at least $d$ independent observations (training samples) are required, to ensure that the covariance matrix of the distribution is not rank deficient.  Obviously, for an infinite dimensional RKHS, this requirement cannot be met. While, in principle, it is possible to regularize the distributions, \eg,~\cite{Zhou_PAMI_2006}, we believe that an in-depth analysis of this approach to kernelize FVs goes beyond the scope of this paper. Note, however, that our approximations of $\mathcal{H}$ can be applied verbatim to derive approximate formulations of kernel FV.

\section{Experiments}
\label{sec:experiments}

We now evaluate our different algorithms, i.e., kVLAD, nVLAD, sVLAD and fVLAD, on several recognition tasks. As mentioned before, 
our main motivation for this work was to be able to exploit the power of the VLAD aggregation scheme to tackle problems where 
the input data is not in vectorial form. Therefore, we focus on two such types of data, which have become increasingly popular in computer vision, namely 
Covariance Descriptors (CovDs), which lie on SPD manifolds, and linear subspaces which form Grassmann manifolds.
Nevertheless, in addition to this manifold-valued data, we also evaluate our algorithms in Euclidean space.

\subsection{SPD Manifold}

In computer vision, SPD matrices have been shown to provide powerful representations for images and videos via region 
covariances~\cite{Tuzel_ECCV_2006}. Such representations have been successfully employed to categorize, e.g., textures~\cite{Tuzel_ECCV_2006,Harandi_ECCV2_2014}, pedestrians~\cite{Tuzel_PAMI_2008} and faces~\cite{Harandi_ECCV2_2014}.

SPD matrices can be thought of as an extension of positive numbers and form the interior of the positive semidefinite cone. It is possible to directly employ the Frobenius norm as a similarity measure between SPD matrices, hence analyzing problems involving such matrices via Euclidean geometry. However, as several studies have shown, undesirable phenomena may occur when Euclidean geometry is utilized to manipulate SPD matrices~\cite{Pennec_IJCV_2006,Tuzel_PAMI_2008,Sadeep:CVPR:2013}. Here, instead, we make use of the Stein divergence defined as
\begin{equation}
\delta_S^2(\Mat{A},\Mat{B}) = \ln\det\Big(\frac{\Mat{A} + \Mat{B}}{2}\Big) -\frac{1}{2}\ln\det\big(\Mat{A}\Mat{B}\big)\;.
\end{equation}
This divergence was shown to yield a positive definite Gaussian kernel~\cite{Sra_NIPS_2012}, named the Stein kernel given by
 $k_\mathcal{S}:\SPD{n} \times \SPD{n} \to \mathbb{R}$ such that $k_\mathcal{S}(\Mat{A},\Mat{B}) = \exp(-\sigma \delta_S^2(\Mat{A},\Mat{B}))$.
In all our experiments on SPD manifolds, the bandwidth of this kernel was determined by cross-validation on the training data. 

A standard approach when dealing with an SPD manifold consists of flattening the manifold using the diffeomorphism $\log:\SPD{n} \to \mathrm{Sym(n)}$, where 
$\log$ and $\mathrm{Sym(n)}$ denote the principal matrix logarithm and the space of symmetric matrices of size $n$, respectively. 
Given that $\mathrm{Sym(n)}$ is a vector space, one can then directly employ tools from Euclidean geometry, here the
VLAD algorithm, to analyze SPD matrices mapped to that space. In our experiments, we refer to this baseline as log-Euclidean VLAD or \emph{lE-VLAD} following the terminology 
used in~\cite{Arsigny_2006}. Note that this strategy has been successfully employed in several recent studies (\eg, for image semantic segmentation~\cite{Carreira_ECCV_2012}). 

Furthermore, we also compare our algorithms against the state-of-the-art Weighted ARray of COvariances (WARCO)~\cite{Tosato_PAMI_2013}, 
{Covariance Discriminative Learning (CDL)~\cite{WANG_CVPR_2012} and Riemannian Sparse Representation using the Stein divergence (RSR-S)~\cite{Harandi_TNNLS_2015} algorithms.}
In WARCO, an image is decomposed into a number of overlapped patches, each of which is represented with a CovD. 
Classification is then performed by combining the output of a set of kernel classifiers trained on local patches. In essence, WARCO pursues the same goal as us, \ie, to aggregate local non-vectorial descriptors, which makes it probably the most relevant baseline, here.
{By contrast, following~\cite{WANG_CVPR_2012,Harandi_TNNLS_2015}, we have used both CDL and RSR-S holistically, \ie, 
every image was described by one SPD matrix.}

In the following experiments on the SPD manifold, we used a codebook of size 32 for all variants of the VLAD algorithm. Empirically, we observed that, for any algorithm, larger codebooks did not significantly improve
the performance. To provide a fair comparison against WARCO, we use the same set of features as~\cite{Tosato_PAMI_2013}. More specifically, from a local patch, a $13 \times 13$ CovD is extracted using the features
\begin{small}
\begin{equation*}
f(x,y) = [h_1(Y),\cdots,h_8(Y), Y, C_b, C_r, \|g(Y)\|, \angle(g(Y))]^T\;,
\end{equation*}
\end{small}
where $f(x,y)$ denotes the feature vector at location $(x,y)$ and $Y$, $C_b$ and $C_r$ 
are the three color channels from the CIELab color space at $(x,y)$. $h_i(\cdot)$ is 
the scaled symmetric Difference Of Offset Gaussian filter bank, and $\|g(Y)\|$ and $\angle(g(Y))$ are the 
gradient magnitude and orientation calculated on the $Y$ channel (see~\cite{Tosato_PAMI_2013} for details).
{The same set of features was used for CDL and RSR-S.}

\paragraph{Head Orientation Classification.}

As a first experiment, we consider the problem of classifying head orientation using the \emph{QMUL} and 
\emph{HOCoffee} datasets~\cite{Tosato_PAMI_2013}. 
The QMUL head dataset contains 19292 images of size $50 \times 50$, captured in an airport terminal. 
The HOCoffee dataset 
(see Fig.~\ref{fig:HOC_example} for examples) 
contains 18117 head images of size $50 \times 50$.
The images typically include a margin of 10 pixels on average, so that the actual average dimension of the heads is
$30 \times 30$ pixels. Both datasets come with predefined training and test samples.

In Table.~\ref{tab:crr_spd}, we report the performance of kVLAD, sVLAD and nVLAD, as well as of  WARCO and lE-VLAD,
on the QMUL and HOCoffee datasets. Note that kVLAD and sVLAD both yield higher accuracies than the state-of-the-art WARCO algorithm.
For example, on HOCoffee, the accuracy of kVLAD surpasses that of WARCO by more than 5\%. 
Note also that kVLAD and sVLAD yield very similar accuracies, which evidences the good quality of our local subspace approximation.
Interestingly, sVLAD even outperforms kVLAD on QMUL. This can be attributed to the square root normalization, which is not possible for kVLAD. Without this normalization, the performance of sVLAD drops by roughly 1\%, and thus remains close to, but slightly lower than that of kVLAD.
Among the approximations, sVLAD is superior to nVLAD. This is not really surprising, since nVLAD relies on a single subspace for all its codewords, whereas sVLAD exploits more local representations.

\def \HOCoffee_SIZE {0.8}
\begin{figure}[!tb]
  \centering
  \includegraphics[width = \HOCoffee_SIZE \columnwidth,keepaspectratio]{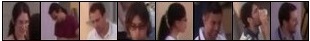}  
  \caption   {  \small    Samples from the HOCoffee dataset.    }
  \label{fig:HOC_example}
\end{figure}
 
\paragraph{Body Orientation Classification.}

As a second task on the SPD manifold, we consider the problem of determining body orientation from images using the Human Orientation Classification (HOC) dataset~\cite{Tosato_PAMI_2013}. The HOC dataset contains 11881 images of size $64 \times 32$  
(see Fig.~\ref{fig:HOCoffee_example} for examples) 
and comprises 4 orientation classes (Front, Back, Left, and Right). In Table.~\ref{tab:crr_spd}, we compare the performance of kVLAD, sVLAD and nVLAD against that of WARCO and lE-VLAD. First, we note that all VLAD variants, including lE-VLAD, are superior to the WARCO algorithm. This demonstrates 
the effectiveness of the VLAD aggregation scheme. Moreover, we note that all our algorithms outperform lE-VLAD. The highest accuracy is obtained by sVLAD which again, in comparison to kVLAD, benefits from the square root normalization.

Altogether, our experiments on SPD manifolds demonstrate that our approach offers an attractive solution to exploiting the information from local patches. Note that, except for a handful of studies (\eg, WARCO), CovDs are usually extracted from entire images, hence making them questionable for challenging classification tasks. This is typically due to the fact that aggregating non-vectorial is an open problem, to which we provide a solution in this paper.

\def \HOC_SIZE {1.0}
\begin{figure}[!tb]
  \centering
  \includegraphics[width = \HOC_SIZE \columnwidth,keepaspectratio]{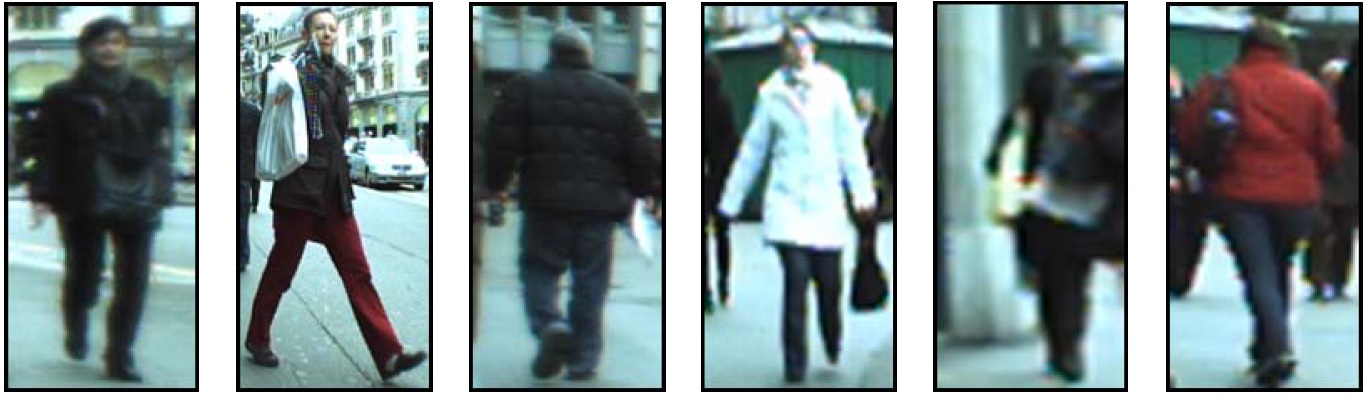}  
  \caption   {  \small    Samples from the HOC dataset.    }
  \label{fig:HOCoffee_example}
\end{figure}

\begin{table}[tb]
	\small
	\centering
  \begin{tabular}{l@{}ccc}
    \toprule
    {\bf Method}   							&{\bf QMUL} 	&{\bf HOCoffee}		&{\bf HOC}		\\
    \toprule    
    \textbf{WARCO}~\cite{Tosato_PAMI_2013} 	& $91\%$		& $80 \%$		& $78 \%$		\\
    {\textbf{CDL}~\cite{WANG_CVPR_2012}} 		& {$81.6\%$}		
    & {$71.2 \%$}		& {$77.8 \%$}		\\
    {\textbf{RSR-S}~\cite{Harandi_TNNLS_2015}}	& {$82.7\%$}		
    & {$65.9 \%$}		& {$78.9 \%$}		\\
    \midrule
    \textbf{lE-VLAD}						& $87.6\%$ 		& $77.2 \%$		& $79.7 \%$\\
    \midrule
	\textbf{nVLAD}         					& $88.9\%$  	& $83.4 \%$		& $81.4 \%$ 	\\  
	\textbf{sVLAD}          				& $92.7\%$  	& $84.0 \%$		& $84.1 \%$ 	\\ 
	\textbf{kVLAD}          				& $92.2\%$  	& $85.3 \%$		& $83.1 \%$		\\ 
	\bottomrule
  \end{tabular}
	\caption    {\small Recognition accuracies for QMUL, HOCoffe and HOC.         }
   \label{tab:crr_spd}
\end{table}

\subsection{Grassmann Manifold}

The space of $p$ dimensional subspaces in $\mathbb{R}^d$ for $0 < p \leq d$ is not a Euclidean space, but a Riemannian manifold known as the Grassmann manifold $\GRASS{p}{d}$. A point $\mathcal{U} \in \GRASS{p}{d}$ is typically represented by a ${d \times p}$ matrix $\Mat{U}$ with orthonormal columns, such that $\mathcal{U} = \mathrm{Span}(\Mat{U})$. The choice of the basis to represent $\mathcal{U}$ is arbitrary and metrics on $\GRASS{p}{d}$ are defined so as to be invariant to this choice. The projection distance is a typical choice of such metric. It was recently shown to induce a valid positive definite kernel on $\GRASS{p}{d}$~\cite{Harandi_ECCV_2014}, \ie, the projection RBF kernel defined as
\begin{equation}
k_p(\Mat{A},\Mat{B}) = \exp(\sigma \|\Mat{A}^T\Mat{B}\|_F^2),~\sigma > 0\;.
\label{eqn:rbf_proj_grass}
\end{equation}
As for the SPD manifold, in our experiments, the bandwidth of this kernel was obtained by cross-validation on the training data.
 
Several state-of-the-art image-set matching methods model sets of images as subspaces~\cite{Harandi_ICCV_2013,Harandi_ECCV_2014}. However, to the best of our knowledge, all these methods rely on a holistic subspace representation. This again is probably due to the fact that, before this paper, no aggregation schemes on Grassmann manifolds have ever been proposed. Our approach, by contrast, enables us to break an image-set into smaller blocks, represent each block by a linear subspace, and aggregating these subspace to form a complete image-set descriptor. 

{
In our experiments, we compare the results of our algorithms against four baselines: First, similarly to the log-Euclidean approach on SPD manifolds, we propose to flatten $\GRASS{p}{d}$ at $\mathbf{I}_{d \times p}$\footnote{We use $\mathbf{I}_{d \times p}$ to denote the truncated identity matrix.} and perform conventional VLAD in the resulting Euclidean space. We refer to this method as lE-VLAD. As a second baseline, we make use of the state-of-the-art Grassmannian Sparse Coding (gSC) algorithm of~\cite{Harandi_ICCV_2013}, which describes each image-set with a single linear subspace. We also employ the kernel version of the Affine Hull Method (kAHM) introduced in~\cite{Cevikalp_CVPR_2010} and the CDL algorithm~\cite{WANG_CVPR_2012} as other state-of-the-art baselines for image-set matching.}
Below, we evaluate the performance of our algorithms and of the baselines on three different classification problems, \ie, object recognition, action classification and pose categorization from image-sets.

\paragraph{Action Recognition.}

As a first experiment on the Grassmannian, we make use of the Ballet dataset~\cite{Ballet_Dataset}. The Ballet dataset consists of 8 complex motion patterns performed by 3 subjects (see Fig.~\ref{fig:ballet_example} for examples).
We extracted 1200 image-sets by grouping 5 frames depicting the same action into one image-set.
The local descriptors for each image-set were obtained by splitting the set into small blocks of size $32 \times 32 \times 3$ and
utilizing Histogram of Oriented Gradient (HOG)~\cite{Dalal:2005:HOG}. We then created subspaces of size $31 \times 3$, hence points on $\GRASS{3}{31}$.
We randomly chose 50\% of imagesets for training and used the remaining sets as test samples. 
The process of random splitting was repeated ten times and the average classification accuracy is reported.

In Table~\ref{tab:crr_grassmann}, we report the accuracy of algorithms and of the gSC and lE-VLAD baselines.
First, note that all the local approaches outperform the holistic gSC method. Furthermore, similarly to the two experiments on SPD manifolds, the maximum accuracy is obtained by sVLAD, closely followed by kVLAD.

{
Given the simplicity of the lE-VLAD method, it is interesting to verify if it can measure up to our kernel extensions by enlarging its dictionary. To this end, we increased the size of the dictionary in lE-VLAD up to the point where the performance started to decrease (256 atoms). While this indeed improved the accuracy of lE-VLAD up to the best accuracy of 91.7\%, it remains significantly below the performance of sVLAD.}

\def \BALLET_SIZE {0.20}
\begin{figure}[!t]
	\centering
    \includegraphics[width= \BALLET_SIZE \columnwidth,keepaspectratio]{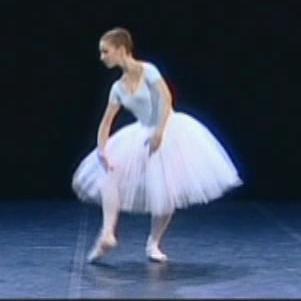}
    \includegraphics[width= \BALLET_SIZE \columnwidth,keepaspectratio]{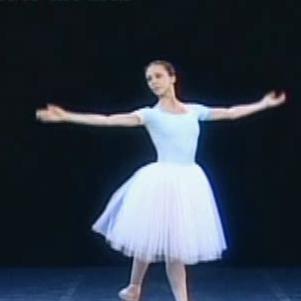}
    \includegraphics[width= \BALLET_SIZE \columnwidth,keepaspectratio]{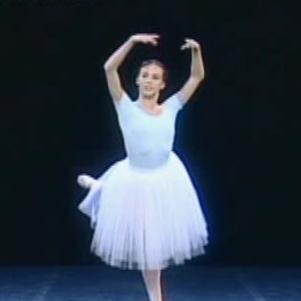}
    \includegraphics[width= \BALLET_SIZE \columnwidth,keepaspectratio]{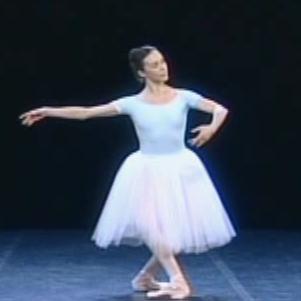}
  \caption{\small Samples from the Ballet dataset.(grayscale images were used in our experiments).}
  \label{fig:ballet_example}
\end{figure}

\paragraph{Object Recognition.}

For the task of object recognition from image-sets, we used the CIFAR dataset~\cite{CIFAR_DB}. The CIFAR dataset contains
60000 images ($32 \times 32$ pixels) from 10 different object categories.  From this dataset, we generated 6000 image-sets, each one containing 10 random images of the same object. In our experiments, we used 1500 image-sets for training and the remaining 4500 image-sets as test data. We report accuracies averaged over 10 random image-set generation processes.

To generate local descriptors, we decomposed each image-set into small blocks of size $8 \times 8 \times 5$. Each block
was then represented by a point on $\GRASS{5}{64}$ using SVD. In Table.~\ref{tab:crr_grassmann}, we compare the results of kVLAD, sVLAD and nVLAD
against those of lE-VLAD and gSC. Here, kVLAD yields the best accuracy followed by sVLAD.

\paragraph{Pose Classification.}

As a last experiment on the Grassmannian, we evaluated the performance of our algorithms on the task of pose categorization using 
the CMU-PIE face dataset~\cite{Sim:PAMI:2003}. The CMU-PIE face dataset contains images of 67 subjects under 13 different poses and 21 different illuminations (see Fig.~\ref{fig:PIE_samples} for examples). 
The images were closely cropped to enclose the face region and resized to $64 \times 64$. 
We extracted 1700 image-sets by grouping 6 images with the same pose, but different illuminations into one image-set.
The local descriptors for each image set were obtained by splitting the set into small blocks of size $32 \times 32 \times 3$ from which we computed Histogram of LBP~\cite{Ojala:PAMI:2002}. We then created subspaces
of size $58 \times 3$, hence points on $\GRASS{3}{58}$. Table~\ref{tab:crr_grassmann} compares the results of nVLAD, sVLAD and kVLAD 
 against those of gSC and lE-VLAD. The highest accuracy is obtained by kVLAD, this time by a large margin over the second best, sVLAD. Note that, with this dataset, flattening the manifold through its tangent space at $\mathbf{I}_{58 \times 3}$ seems to incur strong distortions, as indicated by low performance of lE-VLAD.

\def \PIE_SIZE {0.08}
\begin{figure}[!tb]
      \centering
      \begin{subfigure}[b]{\PIE_SIZE \textwidth}
        	\includegraphics[width=\textwidth]{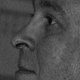}
      \end{subfigure}%
      \begin{subfigure}[b]{\PIE_SIZE \textwidth}
        	\includegraphics[width=\textwidth]{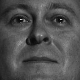}
      \end{subfigure}%
      \begin{subfigure}[b]{\PIE_SIZE \textwidth}
      		\includegraphics[width=\textwidth]{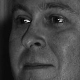}
      \end{subfigure}%
      \begin{subfigure}[b]{\PIE_SIZE \textwidth}
      		\includegraphics[width=\textwidth]{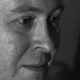}
      \end{subfigure}%
      \begin{subfigure}[b]{\PIE_SIZE \textwidth}
      		\includegraphics[width=\textwidth]{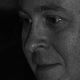}
      \end{subfigure}%
      \begin{subfigure}[b]{\PIE_SIZE \textwidth}
      		\includegraphics[width=\textwidth]{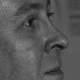}
      \end{subfigure}\\
      \begin{subfigure}[b]{\PIE_SIZE \textwidth}
        	\includegraphics[width=\textwidth]{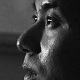}
      \end{subfigure}%
      \begin{subfigure}[b]{\PIE_SIZE \textwidth}
        	\includegraphics[width=\textwidth]{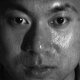}
      \end{subfigure}%
      \begin{subfigure}[b]{\PIE_SIZE \textwidth}
      		\includegraphics[width=\textwidth]{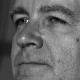}
      \end{subfigure}%
      \begin{subfigure}[b]{\PIE_SIZE \textwidth}
      		\includegraphics[width=\textwidth]{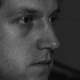}
      \end{subfigure}%
      \begin{subfigure}[b]{\PIE_SIZE \textwidth}
      		\includegraphics[width=\textwidth]{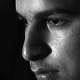}
      \end{subfigure}%
      \begin{subfigure}[b]{\PIE_SIZE \textwidth}
      		\includegraphics[width=\textwidth]{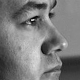}
      \end{subfigure}%
      \caption{Samples from CMU-PIE.}
      \label{fig:PIE_samples}
\end{figure}

\begin{table*}
	\small
	\centering   
  \begin{tabular}{lccc}
    \toprule
    {\bf Method}   			&{\bf Ballet} 	&{\bf CIFAR}		&{\bf CMU-PIE}\\
    \toprule
    \textbf{gSC~\cite{Harandi_ICCV_2013}}			& $79.7 \%$ 		& $59.9 \%$			& $75.5\%$	\\
     {\textbf{kAHM~\cite{Cevikalp_CVPR_2010}}}			& {$85.8\%$} 		
    & {$36.1 \%$}			& {$55.3\%$}	\\
    {\textbf{CDL~\cite{WANG_CVPR_2012}}}			& {$73.1\%$} 		
    & {$54.7 \%$}			& {$64.6\%$}	\\
    \midrule
    \textbf{lE-VLAD }		& $91.1 \%$ 		& $46.2 \%$			& $59.6\%$	\\    
    \midrule
	\textbf{nVLAD}          & $88.9\%$  	& $62.2 \%$ 		& $79.5 \%$\\  
	\textbf{sVLAD}          & $94.4\%$  		& $65.2 \%$			& $80.1\%$\\ 
	\textbf{kVLAD}          & $92.2\%$  	& $67.9 \%$			& $86.3 \%$\\ 
	\bottomrule
  \end{tabular}
  \caption {\small   Accuracies for Ballet, CIFAR and CMU-PIE.}
  \label{tab:crr_grassmann}
\end{table*}

\begin{table*}
	\small
	\centering
\begin{minipage}[t]{.450 \textwidth}
  \small
  \centering   
  \begin{tabular}{lc}
    \toprule
    {\bf Method}   			&{\bf mAP}\\
    \toprule
    {\textbf{SPM}}~\cite{Lazebnik_CVPR_2006}			& {$54.3\%$}\\    
    {\textbf{OCP}}~\cite{Russakovsky_ECCV_2012}			& {$57.2\%$}\\
    {\textbf{Sup-VLAD}}~\cite{Peng_ECCV_2014}			&{$60.9\%$}\\
    \midrule
    \textbf{VLAD}			& $54.7\%$\\
    \midrule
	\textbf{nVLAD}          & $56.2\%$\\  
	\textbf{fVLAD}         	& $55.8\%$\\ 
	\textbf{sVLAD}          & $60.3\%$\\ 
	\bottomrule
  \end{tabular}
  \caption{\small   mean Average Precision (mAP) for VOC 2007 dataset.}
  \label{tab:crr_voc}
  \end{minipage}
  \quad
\begin{minipage}[t]{.450 \textwidth}
  \small
  \centering  
  \begin{tabular}{lc}
    \toprule
    {\bf Method}   			&{\bf CCR}\\
    \toprule
    {\textbf{aLDA}}~\cite{Sharan_IJCV_2013}		& {$44.6\%$}\\
    {\textbf{MS4C}}~\cite{Li_PR_Springer_2014}	& {$50.0\%$}\\
    {$\bf{DTD}_{RBF}$}~\cite{Cimpoi_CVPR_2014}	& {$53.1\%$}\\
    \midrule
    \textbf{VLAD }			& $49.4\%$	\\
    \midrule
	\textbf{nVLAD}          & $52.3\%$ \\  
	\textbf{fVLAD}         	& $50.3\%$ \\ 
	\textbf{sVLAD}          & $55.2\%$	\\ 
	\bottomrule
  \end{tabular}
  \caption{\small   Correct Classification Rate (CCR) for FMD dataset.}
  \label{tab:crr_fmd}
\end{minipage}
\end{table*}

{
\begin{remark}
Several recent studies (\eg,~\cite{WANG_CVPR_2012,Huang_PR_2015}) have tackled the problem of image-set matching using
the geometry of SPD manifolds via covariance descriptors. Table~\ref{tab:crr_grassmann}, however, suggests that, for our experiments, the resulting global covariance descriptors 
do not measure up to subspaces, as evidenced by the performance of CDL in comparison to gSC. We conjecture that this is due
to the small number of images  in each set, which makes the SPD matrices rank deficient (regularization was used to overcome this issue) and less discriminative. 
Interestingly, however, we also evaluated sVLAD using local SPD matrices instead of local subspace, and achieved an accuracy of 93.4\% on the Ballet dataset.
While this remains slightly below what Grassmannian geometry can achieve, it clearly shows the strength of our framework, which, by using a local representation, outperforms the global descriptors of CDL by more than 20\%.
\end{remark}
}

\subsection{Euclidean Space}

Our final experiments are devoted to Euclidean spaces. To this end, we compare the performance of sVLAD, fVLAD and nVLAD against
the conventional VLAD (implementation provided in~\cite{VL_FEAT}) on Pascal VOC 2007~\cite{Everingham:IJCV:2010} and 
on the Flicker Material Database (FMD)~\cite{Sharan_IJCV_2013} (see Fig.~\ref{fig:fmd_example} for examples) . 
Pascal VOC 2007~\cite{Everingham:IJCV:2010} contains 9963 images from 20 object categories.
The FMD contains 1000 images from 10 different material categories~\cite{Sharan_IJCV_2013}. 
Both datasets have been extensively used to benchmark coding techniques.

In our experiments, we realized that the computational load of kVLAD becomes
overwhelming on Pascal VOC07 and FMD as a result of large amount of local descriptors. Hence, 
we will only report the performance of nVLAD, fVLAD and sVLAD here.
The size of codebooks was set to 256 and SIFT descriptors (with whitening) were considered as local features. 
For fVLAD and nVLAD, the size of the RKHS was chosen to be 256 (almost 3 times larger than the original space).
While increasing the dimensionality of the RKHS could potentially improve the results, it would come at the expense of increasing the computational burden of coding.

{
Table~\ref{tab:crr_voc} compares the recognition accuracies of the proposed coding techniques with conventional VLAD, Spatial Pyramid Matching (SPM)~\cite{Lazebnik_CVPR_2006}, Object-Centric spatial Pooling (OCP)~\cite{Russakovsky_ECCV_2012} and supervised 
dictionary learning for VLAD (Sup-VLAD)~\cite{Peng_ECCV_2014}. }
{
Similarly to our experiments on manifolds, sVLAD outperforms the fixed approximation techniques (\ie, fVLAD and nVLAD).
Importantly, we observe that our three algorithms outperform traditional methods such as SPM and VLAD. Furthermore,
sVLAD also outperforms the state-of-the-art pooling method OCP~\cite{Russakovsky_ECCV_2012}, and performs on par with the supervised Sup-VLAD. This latter comparison motivates an interesting future research direction to learn a supervised dictionary in RKHS.
}

{
Table~\ref{tab:crr_fmd} compares the recognition accuracies of nVLAD, fVLAD and sVLAD against VLAD and 
the state-of-the-art methods augmented Latent Dirichlet Allocation (aLDA)~\cite{Sharan_IJCV_2013}, Multi-Scale Spike-and-Slab Sparse Coding (MS4C)~\cite{Li_PR_Springer_2014},
and Describable attributes ($\rm{DTD}_{RBF}$)~\cite{Cimpoi_CVPR_2014} on the FMD dataset.
In essence, we can see that (i) our algorithms outperform VLAD, with sVLAD the best-performing method; (ii) our algorithms outperform 
the state-of-the-art aLDA and MS4C methods; and (iii) while DTD yields higher accuracy than our fixed approximations (\ie, nVLAD and fVLAD),
it is still outperformed by our sVLAD algorithm.
}

\def \FMD_SIZE {0.225}
\begin{figure}[!t]
	\centering
    \includegraphics[width= \FMD_SIZE \columnwidth,keepaspectratio]{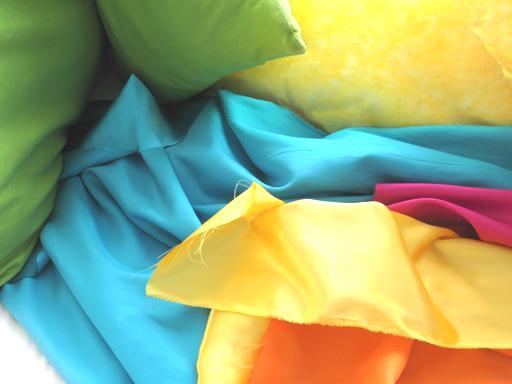}
    \includegraphics[width= \FMD_SIZE \columnwidth,keepaspectratio]{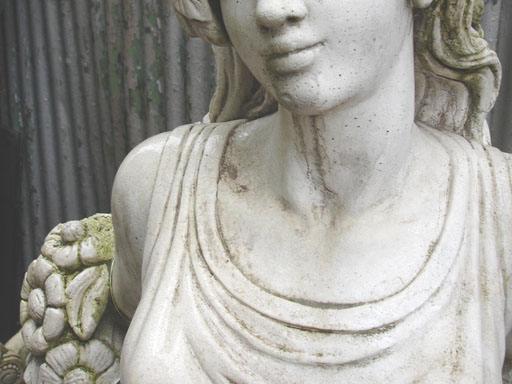}
    \includegraphics[width= \FMD_SIZE \columnwidth,keepaspectratio]{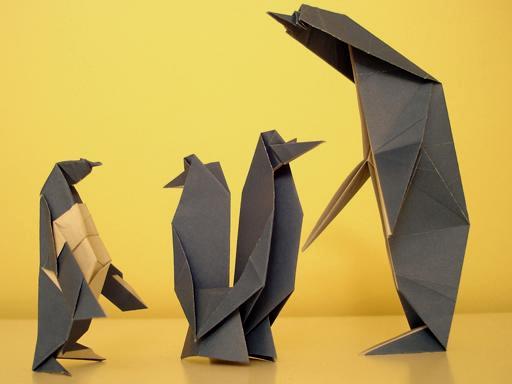}
    \includegraphics[width= \FMD_SIZE \columnwidth,keepaspectratio]{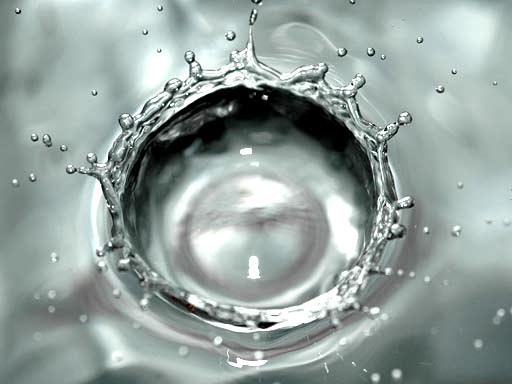}
  \caption{\small Examples of the FMD texture dataset.}
  \label{fig:fmd_example}
\end{figure}

\subsection{Encoding times}

Before concluding, we provide the coding times for the proposed methods on the three different geometries studied
in this work. In particular, we measured the encoding times of sVLAD, fVLAD and nVLAD on a Quad-core machine using Matlab. We also measured 
the running time to compute Eq.~\ref{eqn:nl_vlad_sim_XY}, which shows the computational load of kVLAD. 

The parameters values of the algorithms when measuring these timings were those used in our experiments.
More specifically, for the SPD and Grassmann manifolds, 
the size of codebook was chosen to be 32, while, in the case of Euclidean space, it was set to 256. Note that for the Euclidean case, we assumed that 1000 local descriptors were computed on each image, while, 
for the SPD and Grassmann manifolds, this number was set to 100. 
Table~\ref{tab:running_time} reports all the running times. 

\begin{table}[tb]
	\small
	\centering	
  \begin{tabular}{lccc}
    \toprule
    {\bf Method}   			&{\bf SPD} 	&{\bf Grassmann}	&{\bf Euclidean}\\
    \toprule
 	\textbf{nVLAD}          & 650ms  	& 1600ms  			&35ms\\  
	\textbf{fVLAD}         	& N/A  		&  N/A					&100ms\\ 
	\textbf{sVLAD}          & 750ms  	& 1700ms 				&950ms\\ 
	\textbf{kVLAD}          & 80ms  	& 155ms				&45ms\\
	\bottomrule
  \end{tabular}
  \caption
    {\small    Running times for fVLAD, nVLAD, sVLAD and kVLAD on three different geometries. Note that the running
    times for fVLAD, nVLAD and sVLAD show the coding time for an image/video, while, in the case of kVLAD where not explicit encoding is performed, it shows the
    time needed to evaluate Eq.~\ref{eqn:nl_vlad_sim_XY}
    }
   \label{tab:running_time}     
\end{table}

\section{Conclusions and Future Work}
\label{sec:conclusion}

In this paper, we have introduced a kernel extension of the VLAD encoding scheme. We have also proposed
several approximations to this kernel formulation in the interest of speeding up the encoding process. Not only do the resulting algorithm let us exploit more sophisticated classification schemes in the VLAD framework, but they also allow us to aggregate local descriptors that do not lie in Euclidean space.
Our experiments have evidenced that our algorithms outperform state-of-the-art methods, such as WARCO~\cite{Tosato_PAMI_2013}
and gSC~\cite{Harandi_ICCV_2013}, on several manifold-based recognition tasks. Furthermore, they have also shown that our new encoding schemes yield superior results compared to the conventional VLAD algorithm.
In the future, we plan to explore possible ways of kernelizing the Fisher vector~\cite{Perronnin_CVPR_2007} method. 
We also intend to study the concept of coresets~\cite{Har_2004} to reduce the computational complexity of coding.

{
\bibpunct{(}{)}{;}{a}{,}{,}
\bibliographystyle{natbib}
\bibliography{references}

\begin{thebibliography}{51}
\providecommand{\natexlab}[1]{#1}
\providecommand{\url}[1]{\texttt{#1}}
\expandafter\ifx\csname urlstyle\endcsname\relax
  \providecommand{\doi}[1]{doi: #1}\else
  \providecommand{\doi}{doi: \begingroup \urlstyle{rm}\Url}\fi

\bibitem[Arandjelovic and Zisserman(2013)]{Arandjelovic_CVPR_2013}
Relja Arandjelovic and Andrew Zisserman.
\newblock All about vlad.
\newblock In \emph{CVPR}, 2013.

\bibitem[Aronszajn(1950)]{Aronszajn_1950}
Nachman Aronszajn.
\newblock Theory of reproducing kernels.
\newblock \emph{Transactions of the American mathematical society}, 1950.

\bibitem[Arsigny et~al.(2006)Arsigny, Commowick, Pennec, and
  Ayache]{Arsigny_2006}
Vincent Arsigny, Olivier Commowick, Xavier Pennec, and Nicholas Ayache.
\newblock A log-euclidean framework for statistics on diffeomorphisms.
\newblock In \emph{MICCAI}. 2006.

\bibitem[Bo et~al.(2010)Bo, Ren, and Fox]{Bo_NIPS_2010}
Liefeng Bo, Xiaofeng Ren, and Dieter Fox.
\newblock Kernel descriptors for visual recognition.
\newblock In \emph{NIPS}, 2010.

\bibitem[Carreira et~al.(2012)Carreira, Caseiro, Batista, and
  Sminchisescu]{Carreira_ECCV_2012}
Joao Carreira, Rui Caseiro, Jorge Batista, and Cristian Sminchisescu.
\newblock Semantic segmentation with second-order pooling.
\newblock In \emph{ECCV}. 2012.

\bibitem[Cevikalp and Triggs(2010)]{Cevikalp_CVPR_2010}
Hakan Cevikalp and Bill Triggs.
\newblock Face recognition based on image sets.
\newblock In \emph{CVPR}, pages 2567--2573. IEEE, 2010.

\bibitem[Cimpoi et~al.(2014)Cimpoi, Maji, Kokkinos, Mohamed, and
  Vedaldi]{Cimpoi_CVPR_2014}
Mircea Cimpoi, Subhrajyoti Maji, Iasonas Kokkinos, Salina Mohamed, and Andrea
  Vedaldi.
\newblock Describing textures in the wild.
\newblock In \emph{Computer Vision and Pattern Recognition (CVPR), 2014 IEEE
  Conference on}, pages 3606--3613, 2014.

\bibitem[Dalal and Triggs(2005)]{Dalal:2005:HOG}
Navneet Dalal and Bill Triggs.
\newblock Histograms of oriented gradients for human detection.
\newblock In \emph{CVPR}, 2005.

\bibitem[Everingham et~al.(2010)Everingham, Van~Gool, Williams, Winn, and
  Zisserman]{Everingham:IJCV:2010}
Mark Everingham, Luc Van~Gool, Christopher~KI Williams, John Winn, and Andrew
  Zisserman.
\newblock The pascal visual object classes (voc) challenge.
\newblock \emph{IJCV}, 88\penalty0 (2), 2010.

\bibitem[Gehler and Nowozin(2009)]{Gehler_CVPR_2009}
Peter Gehler and Sebastian Nowozin.
\newblock On feature combination for multiclass object classification.
\newblock In \emph{CVPR}, 2009.

\bibitem[Gong et~al.(2014)Gong, Wang, Guo, and Lazebnik]{Gong_ECCV_2014}
Yunchao Gong, Liwei Wang, Ruiqi Guo, and Svetlana Lazebnik.
\newblock Multi-scale orderless pooling of deep convolutional activation
  features.
\newblock In \emph{ECCV}. 2014.

\bibitem[Grauman and Darrell(2005)]{Grauman:ICCV:2005}
Kristen Grauman and Trevor Darrell.
\newblock The pyramid match kernel: Discriminative classification with sets of
  image features.
\newblock In \emph{ICCV}, 2005.

\bibitem[Har-Peled and Mazumdar(2004)]{Har_2004}
Sariel Har-Peled and Soham Mazumdar.
\newblock On coresets for k-means and k-median clustering.
\newblock In \emph{ACM symposium on Theory of computing}, 2004.

\bibitem[Harandi et~al.(2013)Harandi, Sanderson, Shen, and
  Lovell]{Harandi_ICCV_2013}
Mehrtash Harandi, Conrad Sanderson, Chunhua Shen, and Brian Lovell.
\newblock Dictionary learning and sparse coding on grassmann manifolds: An
  extrinsic solution.
\newblock In \emph{ICCV}, 2013.

\bibitem[Harandi et~al.(2014{\natexlab{a}})Harandi, Salzmann, and
  Hartley]{Harandi_ECCV2_2014}
MehrtashT. Harandi, Mathieu Salzmann, and Richard Hartley.
\newblock From manifold to manifold: Geometry-aware dimensionality reduction
  for spd matrices.
\newblock In \emph{ECCV}. 2014{\natexlab{a}}.

\bibitem[Harandi et~al.(2014{\natexlab{b}})Harandi, Salzmann, Jayasumana,
  Hartley, and Li]{Harandi_ECCV_2014}
MehrtashT. Harandi, Mathieu Salzmann, Sadeep Jayasumana, Richard Hartley, and
  Hongdong Li.
\newblock Expanding the family of grassmannian kernels: An embedding
  perspective.
\newblock In \emph{ECCV}. 2014{\natexlab{b}}.

\bibitem[Harandi et~al.(2015)Harandi, Hartley, Lovell, and
  Sanderson]{Harandi_TNNLS_2015}
M.T. Harandi, R.~Hartley, B.~Lovell, and C.~Sanderson.
\newblock Sparse coding on symmetric positive definite manifolds using bregman
  divergences.
\newblock \emph{TNNLS}, PP\penalty0 (99):\penalty0 1--1, 2015.
\newblock ISSN 2162-237X.
\newblock \doi{10.1109/TNNLS.2014.2387383}.

\bibitem[Huang et~al.(2015)Huang, Wang, Shan, and Chen]{Huang_PR_2015}
Zhiwu Huang, Ruiping Wang, Shiguang Shan, and Xilin Chen.
\newblock Face recognition on large-scale video in the wild with hybrid
  euclidean-and-riemannian metric learning.
\newblock \emph{Pattern Recognition}, 48\penalty0 (10):\penalty0 3113 -- 3124,
  2015.
\newblock ISSN 0031-3203.
\newblock Discriminative Feature Learning from Big Data for Visual Recognition.

\bibitem[Jaakkola et~al.(1999)Jaakkola, Haussler, et~al.]{Jaakkola_NIPS_1999}
Tommi Jaakkola, David Haussler, et~al.
\newblock Exploiting generative models in discriminative classifiers.
\newblock In \emph{NIPS}, 1999.

\bibitem[Jayasumana et~al.(2013)Jayasumana, Hartley, Salzmann, Li, and
  Harandi]{Sadeep:CVPR:2013}
Sadeep Jayasumana, Richard Hartley, Mathieu Salzmann, Hongdong Li, and Mehrtash
  Harandi.
\newblock Kernel methods on the riemannian manifold of symmetric positive
  definite matrices.
\newblock In \emph{CVPR}, 2013.

\bibitem[J{\'e}gou et~al.(2010)J{\'e}gou, Douze, Schmid, and
  P{\'e}rez]{Jegou_CVPR_2010}
Herv{\'e} J{\'e}gou, Matthijs Douze, Cordelia Schmid, and Patrick P{\'e}rez.
\newblock Aggregating local descriptors into a compact image representation.
\newblock In \emph{CVPR}, 2010.

\bibitem[Krizhevsky and Hinton(2009)]{CIFAR_DB}
Alex Krizhevsky and Geoffrey Hinton.
\newblock Learning multiple layers of features from tiny images.
\newblock \emph{Tech. Rep}, 2009.

\bibitem[Krizhevsky et~al.(2012)Krizhevsky, Sutskever, and
  Hinton]{Krizhevsky_NIPS_2012}
Alex Krizhevsky, Ilya Sutskever, and Geoffrey~E Hinton.
\newblock Imagenet classification with deep convolutional neural networks.
\newblock In \emph{NIPS}, 2012.

\bibitem[Lazebnik et~al.(2006)Lazebnik, Schmid, and Ponce]{Lazebnik_CVPR_2006}
Svetlana Lazebnik, Cordelia Schmid, and Jean Ponce.
\newblock Beyond bags of features: Spatial pyramid matching for recognizing
  natural scene categories.
\newblock In \emph{CVPR}, 2006.

\bibitem[Li(2014)]{Li_PR_Springer_2014}
Wenbin Li.
\newblock Learning multi-scale representations for material classification.
\newblock In Xiaoyi Jiang, Joachim Hornegger, and Reinhard Koch, editors,
  \emph{Pattern Recognition}, volume 8753, pages 757--764. Springer
  International Publishing, 2014.

\bibitem[Mairal et~al.(2014)Mairal, Koniusz, Harchaoui, and
  Schmid]{Mairal_NIPS_2014}
Julien Mairal, Piotr Koniusz, Zaid Harchaoui, and Cordelia Schmid.
\newblock Convolutional kernel networks.
\newblock In \emph{NIPS}. 2014.

\bibitem[Ojala et~al.(2002)Ojala, Pietikainen, and Maenpaa]{Ojala:PAMI:2002}
Timo Ojala, Matti Pietikainen, and Topi Maenpaa.
\newblock Multiresolution gray-scale and rotation invariant texture
  classification with local binary patterns.
\newblock \emph{TPAMI}, 24\penalty0 (7), 2002.

\bibitem[Peng et~al.(2014)Peng, Wang, Qiao, and Peng]{Peng_ECCV_2014}
Xiaojiang Peng, Limin Wang, Yu~Qiao, and Qiang Peng.
\newblock Boosting vlad with supervised dictionary learning and high-order
  statistics.
\newblock In \emph{ECCV}. 2014.

\bibitem[Pennec et~al.(2006)Pennec, Fillard, and Ayache]{Pennec_IJCV_2006}
Xavier Pennec, Pierre Fillard, and Nicholas Ayache.
\newblock A riemannian framework for tensor computing.
\newblock \emph{IJCV}, 66\penalty0 (1), 2006.

\bibitem[Perronnin and Dance(2007)]{Perronnin_CVPR_2007}
F.~Perronnin and C.~Dance.
\newblock Fisher kernels on visual vocabularies for image categorization.
\newblock In \emph{CVPR}, 2007.

\bibitem[Perronnin et~al.(2010{\natexlab{a}})Perronnin, S{\'a}nchez, and
  Liu]{Perronnin_CVPR_2010}
Florent Perronnin, Jorge S{\'a}nchez, and Yan Liu.
\newblock Large-scale image categorization with explicit data embedding.
\newblock In \emph{CVPR}, 2010{\natexlab{a}}.

\bibitem[Perronnin et~al.(2010{\natexlab{b}})Perronnin, S{\'a}nchez, and
  Mensink]{Perronnin_ECCV_2010}
Florent Perronnin, Jorge S{\'a}nchez, and Thomas Mensink.
\newblock Improving the fisher kernel for large-scale image classification.
\newblock In \emph{ECCV}. 2010{\natexlab{b}}.

\bibitem[Rahimi and Recht(2007)]{Rahimi_NIPS_2007}
Ali Rahimi and Benjamin Recht.
\newblock Random features for large-scale kernel machines.
\newblock In \emph{NIPS}, 2007.

\bibitem[Rudin(2011)]{Rudin_2011}
Walter Rudin.
\newblock \emph{Fourier analysis on groups}.
\newblock John Wiley \& Sons, 2011.

\bibitem[Russakovsky et~al.(2012)Russakovsky, Lin, Yu, and
  Fei-Fei]{Russakovsky_ECCV_2012}
Olga Russakovsky, Yuanqing Lin, Kai Yu, and Li~Fei-Fei.
\newblock Object-centric spatial pooling for image classification.
\newblock In \emph{ECCV}, pages 1--15. Springer, 2012.

\bibitem[Sharan et~al.(2013)Sharan, Liu, Rosenholtz, and
  Adelson]{Sharan_IJCV_2013}
Lavanya Sharan, Ce~Liu, Ruth Rosenholtz, and Edward~H Adelson.
\newblock Recognizing materials using perceptually inspired features.
\newblock \emph{IJCV}, 103\penalty0 (3), 2013.

\bibitem[Sim et~al.(2003)Sim, Baker, and Bsat]{Sim:PAMI:2003}
Terence Sim, Simon Baker, and Maan Bsat.
\newblock The cmu pose, illumination, and expression database.
\newblock \emph{TPAMI}, 25\penalty0 (12), 2003.

\bibitem[Sivic et~al.(2005)Sivic, Russell, Efros, Zisserman, and
  Freeman]{Sivic_ICCV_2005}
Josef Sivic, Bryan~C Russell, Alexei~A Efros, Andrew Zisserman, and William~T
  Freeman.
\newblock Discovering objects and their location in images.
\newblock In \emph{ICCV}, 2005.

\bibitem[Sra(2012)]{Sra_NIPS_2012}
Suvrit Sra.
\newblock A new metric on the manifold of kernel matrices with application to
  matrix geometric means.
\newblock In \emph{NIPS}, pages 144--152, 2012.

\bibitem[Tosato et~al.(2013)Tosato, Spera, Cristani, and
  Murino]{Tosato_PAMI_2013}
Diego Tosato, Mauro Spera, Marco Cristani, and Vittorio Murino.
\newblock Characterizing humans on riemannian manifolds.
\newblock \emph{TPAMI}, 35\penalty0 (8), 2013.

\bibitem[Tuzel et~al.(2006)Tuzel, Porikli, and Meer]{Tuzel_ECCV_2006}
Oncel Tuzel, Fatih Porikli, and Peter Meer.
\newblock Region covariance: A fast descriptor for detection and
  classification.
\newblock In \emph{ECCV}. 2006.

\bibitem[Tuzel et~al.(2008)Tuzel, Porikli, and Meer]{Tuzel_PAMI_2008}
Oncel Tuzel, Fatih Porikli, and Peter Meer.
\newblock Pedestrian detection via classification on {R}iemannian manifolds.
\newblock \emph{TPAMI}, 30\penalty0 (10), 2008.

\bibitem[van Gemert et~al.(2010)van Gemert, Veenman, Smeulders, and
  Geusebroek]{Van_PAMI_2010}
Jan~C van Gemert, Cor~J Veenman, Arnold~WM Smeulders, and J-M Geusebroek.
\newblock Visual word ambiguity.
\newblock \emph{TPAMI}, 32\penalty0 (7), 2010.

\bibitem[Vedaldi and Fulkerson(2008)]{VL_FEAT}
A.~Vedaldi and B.~Fulkerson.
\newblock {VLFeat}: An open and portable library of computer vision algorithms,
  2008.

\bibitem[Vedaldi and Zisserman(2012{\natexlab{a}})]{Vedaldi_TPAMI_2012}
A.~Vedaldi and A.~Zisserman.
\newblock Efficient additive kernels via explicit feature maps.
\newblock \emph{TPAMI}, 34\penalty0 (3), 2012{\natexlab{a}}.

\bibitem[Vedaldi and Zisserman(2012{\natexlab{b}})]{Vedaldi_PAMI_2012}
Andrea Vedaldi and Andrew Zisserman.
\newblock Efficient additive kernels via explicit feature maps.
\newblock \emph{TPAMI}, 34\penalty0 (3), 2012{\natexlab{b}}.

\bibitem[Wang et~al.(2012)Wang, Guo, Davis, and Dai]{WANG_CVPR_2012}
Ruiping Wang, Huimin Guo, L.S. Davis, and Qionghai Dai.
\newblock Covariance discriminative learning: A natural and efficient approach
  to image set classification.
\newblock In \emph{CVPR}, pages 2496--2503, June 2012.

\bibitem[Wang and Mori(2009)]{Ballet_Dataset}
Yang Wang and Greg Mori.
\newblock Human action recognition by semilatent topic models.
\newblock \emph{TPAMI}, 31\penalty0 (10), 2009.

\bibitem[Winn et~al.(2005)Winn, Criminisi, and Minka]{Winn_ICCV_2005}
John Winn, Antonio Criminisi, and Thomas Minka.
\newblock Object categorization by learned universal visual dictionary.
\newblock In \emph{ICCV}, 2005.

\bibitem[Yang et~al.(2009)Yang, Yu, Gong, and Huang]{Yang_CVPR_2009}
Jianchao Yang, Kai Yu, Yihong Gong, and Thomas Huang.
\newblock Linear spatial pyramid matching using sparse coding for image
  classification.
\newblock In \emph{CVPR}, 2009.

\bibitem[Zhou and Chellappa(2006)]{Zhou_PAMI_2006}
Shaohua~Kevin Zhou and Rama Chellappa.
\newblock From sample similarity to ensemble similarity: Probabilistic distance
  measures in reproducing kernel hilbert space.
\newblock \emph{TPAMI}, 28\penalty0 (6), 2006.

\end{thebibliography}
}

\end{document}